\documentclass{article}
\usepackage{ijcai26}

\usepackage{times}
\usepackage{soul}
\usepackage{url}
\usepackage[hidelinks]{hyperref}
\usepackage[utf8]{inputenc}
\usepackage[small]{caption}
\usepackage{graphicx}
\usepackage{amsmath}
\usepackage{amsthm}
\usepackage{amssymb}
\usepackage{booktabs}
\usepackage{algorithm}
\usepackage{algpseudocode}
 \usepackage{multirow}
\usepackage[switch]{lineno}
\usepackage{graphicx}
\usepackage{subcaption} 

\DeclareMathOperator*{\argmax}{arg\,max}

\usepackage{listings}
\usepackage{xcolor}

\definecolor{codegreen}{rgb}{0,0.6,0}
\definecolor{codegray}{rgb}{0.5,0.5,0.5}
\definecolor{codepurple}{rgb}{0.58,0,0.82}
\definecolor{backcolour}{rgb}{0.95,0.95,0.92}

\lstdefinestyle{promptstyle}{
    backgroundcolor=\color{backcolour},   
    commentstyle=\color{codegreen},
    keywordstyle=\color{magenta},
    numberstyle=\tiny\color{codegray},
    stringstyle=\color{codepurple},
    basicstyle=\ttfamily\footnotesize, 
    breakatwhitespace=false,         
    breaklines=true,                 
    captionpos=b,                    
    keepspaces=true,                 
    numbers=left,                    
    numbersep=5pt,                  
    showspaces=false,                
    showstringspaces=false,
    showtabs=false,                  
    tabsize=2,
    frame=single,                    
    rulecolor=\color{black}
}
\title{GUI agent: Guided Exploration of User-Sensitive Screens 
\thanks{Presented at the IJCAI-ECAI 2026 RobustifAI workshop}}

\author{
Aradhana Nayak $^1$
\and
Mussadiq Nazeer$^1$\and
Wang Peng $^{2}$\And
Feng Liu$^1$\\
\affiliations
$^1$Huawei Heisenberg Research Center (Munich)\\
$^2$Huawei Technologies Co., Ltd
\emails
\{aradhana.nayak, mussadiq.mussadiq, wangpeng394, feng.liu1\}@huawei.com
}

\begin{document}

\maketitle
\begin{abstract}
    LLM agents are increasingly being used to automate tasks for users within an open GUI environment. They inevitably encounter screens containing user-sensitive information, for which takeover of task execution by the user is highly desirable or even necessary. State-of-the-art LLM-driven agents are usually fine-tuned to complete tasks regardless of the safety implications of their actions. This makes their real-world deployment difficult and adversely affects the reliability. Therefore, it is crucial to identify and categorize user-sensitive states and define user-sensitive queries. This dataset would be to engineers to recognize and request handover to the user in critical scenarios. This short paper develops an explorer agent that systematically explores the query space starting from one demonstrated task to identify queries that, if executed, would lead to user-sensitive states in a GUI environment.
\end{abstract}

\section{Introduction}

The rapid development of the thinking and reasoning capabilities of LLMs, combined with their ability to match the performance of older models with an increasingly smaller number of parameters, has led to their widespread application to edge devices (\cite{Mobileagentv3}, \cite{uitars}, \cite{mobilegpt}).  LLMs can perform a wide range of tasks on web and GUI interfaces, from simple tasks such as adding calendar events to long-term tasks such as shopping for items and cross-application tasks. This has necessitated the endowment of LLM-based GUI agents with a high degree of autonomy, which may sometimes infringe users' privacy and conflict with their preferences (\cite{zhang2024privacyasst}, \cite{he2025emerged}). Furthermore, the actions executed by a GUI agent are often irreversible, such as sending emails, deleting files and completing transactions (\cite{hua2024trustagent}). The workflow of GUI agents can easily become uncontrollable, as they operate in closed loops where each action modifies the environment, and a single incorrect click can lead to unexpected screens and cascading errors (\cite{wu2024dissecting}). To safely integrate LLMs as decision makers in GUI environments, it is therefore necessary to identify and isolate screens that could potentially require user feedback. 

\section{Related Work}
\textbf{Exploration in GUI Environment}
Recent works address GUI agent generalization by leveraging exploration-based priors. GUI-Xplore \cite{Guiexplore} uses pre-recorded videos to build transition graphs for cross-app generalization, while GUI-Explorer \cite{Guiexplorer} constructs unsupervised knowledge graphs for dynamic guidance. Guardian \cite{Guardian} offloads computation to external engines to refine action spaces. 
\flushleft\textbf{Reinforcement Learning (RL) methods for LLM agents}
RL enables active, autonomous exploration, reducing reliance on static datasets. This allows agents to dynamically adapt to novel GUI environments. Their application however, is challenging as sparse rewards in dynamic interfaces often fail to provide sufficient guidance for decision making. \cite{enhancing} and \cite{AgentQ} integrate Monte Carlo Tree Search (MCTS) with RL to decompose sparse rewards into dense Q-values, addressing credit assignment; \cite{AgentQ} further uses self-critique as intermediate feedback. \cite{Qclass} propagates future utility backward to prune inefficient behaviors, while Search-Agent \cite{Searchagent} employs A* search with alternative paths for explicit back-tracking. \cite{GuiBee} treats GUI elements as tokens for robust visual grounding, and \cite{MCDML} uses dynamic memory-guided MCTS to adjust action evaluation. \cite{Lats} adapts MCTS via Language Agent Tree Search, expanding ReAct into a reasoning search. 
\flushleft\textbf{MCTS-based Exploration for GUI Environments}
While policy improvement through RL-based methods has received significant attention in literature, their use in dataset curation and out-of-domain generalization remains less explored. By optimizing exploration policies iteratively, RL-based methods could potentially prioritize unvisited states, significantly improving screen diversity coverage and to selectively design training datasets for application specific purposes. While reasoning capabilities of current Reasoning LLMs (RLLMs) are generally desirable for systematic problem solving, \cite{ReasoningLLMsare} challenges this assumption. The authors argue that RLLMs often engage in ``structureless wandering" rather than rigorous exploration. The lack of rigorous exploration of the solution space leads to exponential performance deterioration as problem complexity increases, suggesting that mere computational scaling is insufficient. The solution-space exploration policies in literature are either through instruction-tuning (\cite{Nnetnav}) or purely algorithmic (\cite{Expel}). \cite{ScreenExplorer} propose ScreenExplorer to maximize exploration efficiency through a reward shaping mechanism. It prioritizes all unvisited screen states to discover novel interaction paths with the aim of eventually finding a solution trajectory to every possible user query. However, the performance of the model after several rounds of training is comparable to an untrained SOTA model and, a coverage argument for exploration of solution-space is not provided.

In this short paper, we propose an algorithm to isolate user-sensitive query (or tasks) and screens starting from a single user-defined trajectory in a GUI application. The method is based on MCTS and relies on experience distillation from previous iterations (training rounds) to generate novel user-query. We obtain evidences of coverage of query space 1) within a single training round through instruction finetuning and a proposed saturation algorithm 2) within consecutive rounds through rewards employed in the RL training. 

\begin{figure}
    \centering
    \includegraphics[width=\linewidth, scale=1.5]{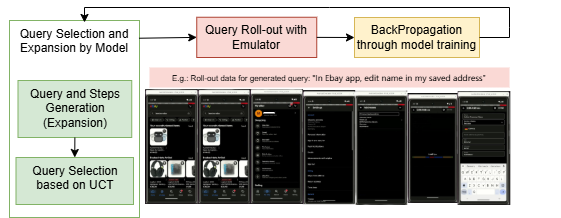}
    \caption{The explorer LM is trained with an MCTS-like approach. The native LM (M3A \protect\cite{Androidworld}) proceeds to modify user-sensitive data during roll-out.}
    \label{fig:framework}
\end{figure}

\section{Methodology}
Our framework consists of two models: the native language (or vision language) model (LM), which determines actions, and the explorer LM, which is the main contribution of this work. The explorer agent is iteratively trained with the objective of pure exploration, i.e. the policy determines which node to explore with rewards based on novelty and user sensitivity of the node. Our exploration objective is dual as the search space comprises both the base query and the sensitive nodes branching out from it. 

\begin{algorithm}
\caption{Learning Framework for Explorer LM}
\label{alg:learning_framework}
\begin{algorithmic}[1]
\Require Exploration model $\Pi_\theta$, Algorithms $\mathcal{A}_{sel}$ (Query Selection), $\mathcal{A}_{sat}$ (Saturation Check), Minimum number of novel queries per iteration $N_{min}$
\State Generate and Select $\mathcal{Q} \sim \Pi_\theta$ with $\mathcal{A}_{sat}$, $\mathcal{A}_{sel}$
\While{$len(Q) >N_{min}$}:
    \State Perform roll-out for all queries in $\mathcal{Q}$
    \State Back propagate gradients from roll-out queries
    \State Update $\Pi_\theta$
    \State Generate and Select $\mathcal{Q} \sim \Pi_\theta$ with $\mathcal{A}_{sat}$, $\mathcal{A}_{sel}$
\EndWhile    
\end{algorithmic}
\end{algorithm}

\subsection{Query Selection, Expansion and Saturation}
In order to select a base query in the query search space, we train the policy $\Pi_\theta(s_0)$ where $s_0$ is the home page screen in the GUI environment. The policy function is the log-likelihood of the output sequence of the explorer LM under it’s distribution. The explorer LM generates $N\times M$ episodes (queries and their steps) in $N$ batches of size $M$ each up to saturation (Algorithm \ref{alg:sat_check}). After generation of queries, selection method (in Algorithm \ref{alg:query_selection}) is used which relies on a single execution trajectory $\mathcal{T}_0 = (u_0, s_{0j}, a_{0j} ); j=1\dotsc T$ of $T$ screens as input along with an instruction $\mathcal{I}$ (details in Appendix). Essentially we simplify the following UCT formula as $N(u_i)=0$ i.e. repeated queries are dropped by Algorithm \ref{alg:query_selection}.
\begin{align*}
    u^* &= \argmax_{i \in \{ 1, \dotsc, M\}} \left( V(u_i, s_0) + \sqrt{\frac{\log N(u_i)}{1 + \log N(u_i)}} \right)\\
    &= \argmax_{i \in \{ 1, \dotsc, M\}}  V(u_i, s_0) 
\end{align*}
where $V(u_i, s_0) := \Pi_{k=1}^M \Pi_{p=1}^P \{Sim(u_i, u_k)-\tau) * Sim(u_i, u_p)-\tau \}$ and $Sim(u_i, u_j)$ is defined in Algorithm \ref{alg:query_selection} ($\mathcal{A}_{sel}$). Instead of a single query $u^*$ in each iteration of MCTS loop, we select a batch of queries $\{u_i\}_{i \in \text{index\_retained}}$ in order to prevent overfitting during training of the explorer LM. This is also the reason we choose GRPO for training the LLM and experience distillation. The selected batch of queries is:
\begin{align*}
    u^* \in \{u_i : V(u_i, s_0) > 0\}
\end{align*}
$\mathcal{A}_{sel}$ enforces that $\text{indices\_retained} = \{i:V(u_i, s_0) > 0\}$ wherein all generated queries in a batch are compared with themselves as well as the pool (initialized with $\mathcal{T}_0$ and pool $P=1$). Queries retained in each batch are added to $P$. The number of batches $N$ is decided according to a Algorithm \ref{alg:sat_check} which identifies the inflection point after which few new queries are generated by the model. Each query $u_i$ is generated along with steps. In particular, actions taken called ``actionPurpose" $a_{ij}$, screen descriptions called ``uiSummary" and ``category" of screens for steps $j \in \{ 0, \dotsc, T-1 \}$ are generated. The number of unique queries is less than or equal to $N \times M$ and the number of steps per query $T$ is dependent on the query itself.

\begin{algorithm}
\caption{Query Selection Algorithm ($\mathcal{A}_{sel}$)}
\label{alg:query_selection}
\begin{algorithmic}[1]
\Require Model $\Pi_\theta$, Instruction $\mathcal{I}$, Pool Size $P$, Threshold $\tau$, execution trajectory $\mathcal{T}_0 = (u_0, s_{0j}, a_{0j} ); j=1\dotsc T$
\Ensure Vectorized Pool $\{u_{p, \text{vec}}\}_{p=1}^P$

\For{$i = 1$ \textbf{to} $M$}
    \State Generate query (and steps): $u_i \sim \Pi_\theta(\cdot | \mathcal{I}, \mathcal{T}_0)$
\EndFor

\State \textbf{Preprocessing:}
\State Compute vectorized representations for all queries:
$u_{i, \text{vec}} \leftarrow \text{Embed}(u_i)$ for each $u_i \in \{u_1, \dots, u_M\}$\;

\State $\text{index\_retained} \leftarrow \emptyset$\; $\text{query\_retained} \leftarrow \emptyset$\;

\For{$i = 1$ \textbf{to} $M$}
    \For{$k =1$ \textbf{to} $i$}
        \State Compute similarity of query in batch to query in batch:
              $Sim(u_i, u_k) \leftarrow \text{CosineSimilarity}(u_{i, \text{vec}}, u_{k, \text{vec}})$\;
        \For{$p = 1$ \textbf{to} $P$}
            \State Compute similarity of query in batch to query in pool:
            $Sim(u_i, u_p) \leftarrow \text{CosineSimilarity}(u_{i, \text{vec}}, u_{p, \text{vec}})$\;
            \If{$Sim(u_i, u_p) < \tau$ $\And$  $Sim(u_i, u_k) < \tau$}
                \State \textbf{Add} $i$ to $\text{index\_retained}$\;
                \State \textbf{Add} $u_i$ to $\text{query\_retained}$\;
            \EndIf
        \EndFor
    \EndFor
\EndFor

\State \Return $\text{index\_retained}$,  $\text{query\_retained}$ 
\end{algorithmic}
\end{algorithm}

\subsection{Roll-out with Emulator}
Our set-up for roll-out is that of SPABench (\cite{Spabench}). The agent is connected to an Android emulator providing the screen observation to the agent. The actions (generated by the agent) are executed within the emulator which outputs a screenshot after each action. The execution trajectory containing screenshots (from emulator) and action logs (from agent) are saved in the worker machine which manages communication between emulator and agent and, oversees the task completion process. The native LM can be freely chosen in SPABench such as SeeAct (\cite{SeeAct}), M3A (\cite{Androidworld}), MobileAgent (\cite{Mobileagent}, \cite{Mobileagentv2}) etc.. We choose M3A agent for the experiments in Section \ref{sec:experiments}.
\begin{figure}
    \centering
    \includegraphics[width=\linewidth, scale=0.23]{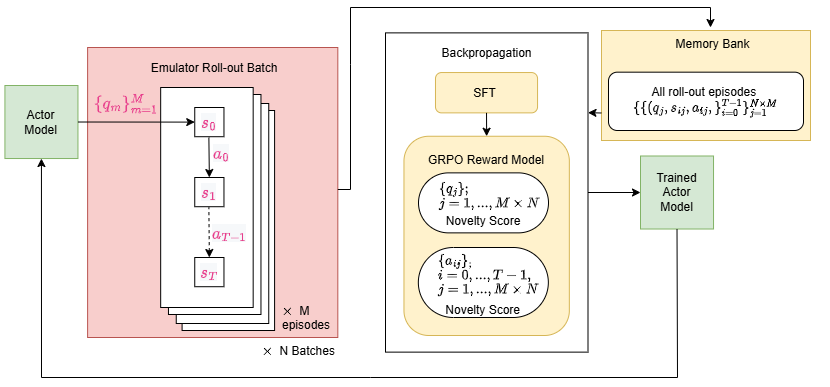}
    \caption{Backpropagation step in MCTS is replaced by GRPO training over a batch of selected queries for Explorer LM}
    \label{fig:placeholder}
\end{figure}
\subsection{Training}\label{sec:training}
\begin{figure}
    \centering
    \begin{subfigure}[b]{0.15\textwidth} 
        \centering
        \includegraphics[trim=0cm 0cm 1.5cm 0cm, clip, width=1.1\linewidth]{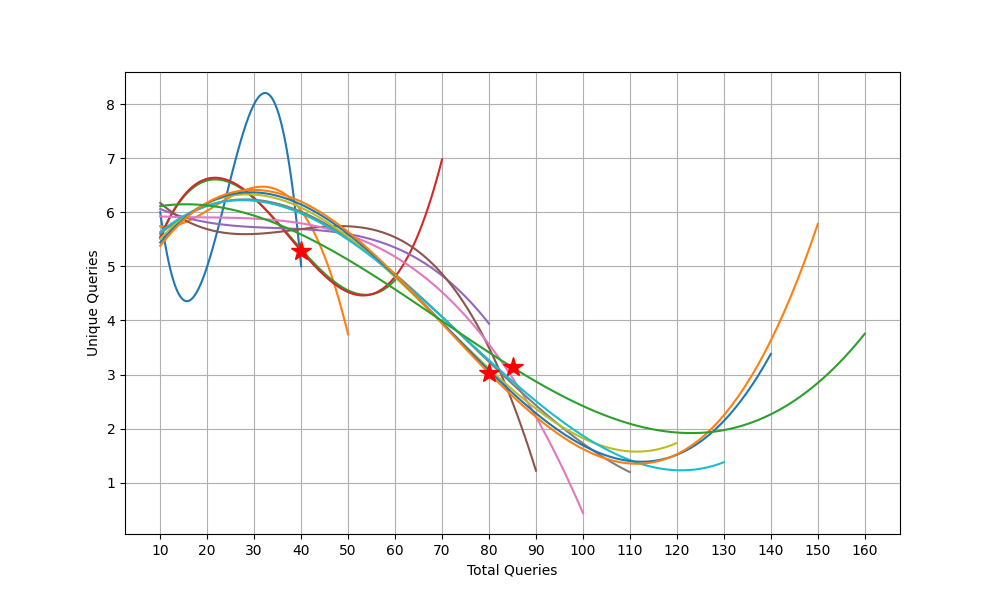}
        \label{fig:sat_round_1}
    \end{subfigure}
    \hfill 
    \begin{subfigure}[b]{0.15\textwidth}
        \centering
        \includegraphics[trim=0cm 0cm 1.5cm 0cm, clip, width=1.1\linewidth]{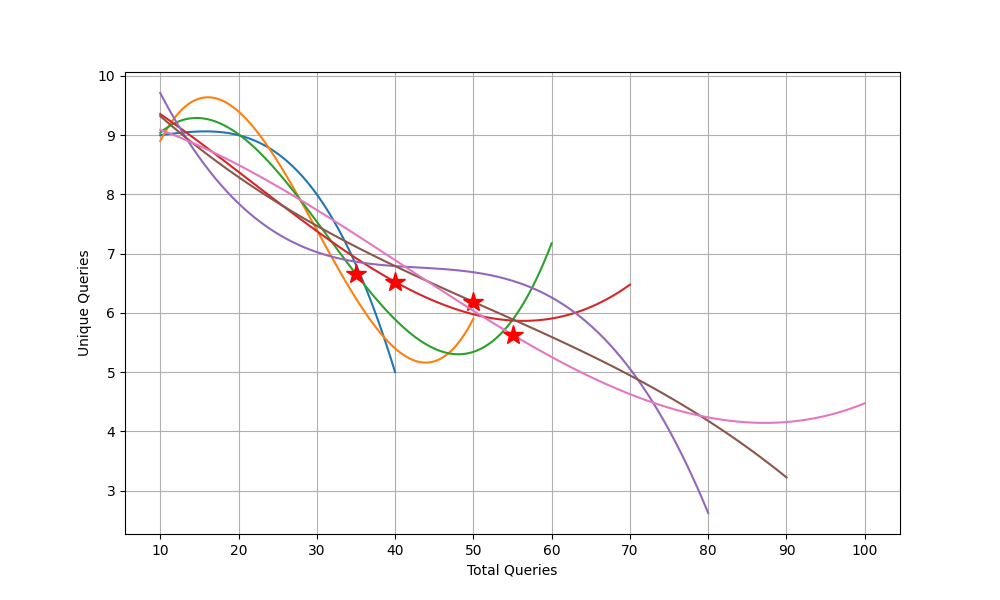}
        \label{fig:sat_round_2}
    \end{subfigure}
    \begin{subfigure}[b]{0.15\textwidth}
        \centering
        \includegraphics[trim=0cm 0cm 1.5cm 0cm, clip, width=1.1\linewidth]{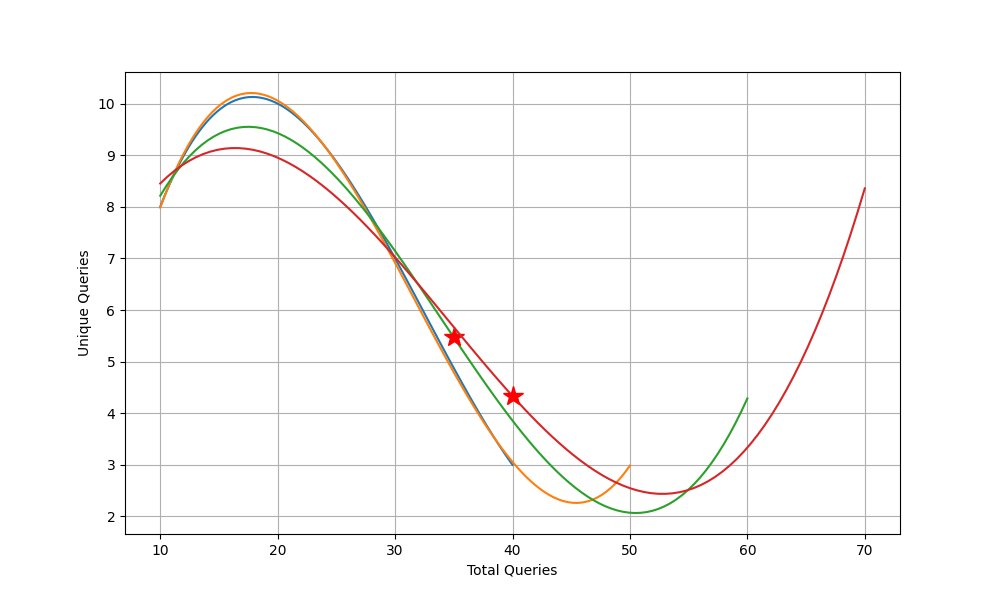}
        \label{fig:sat_round_3}
    \end{subfigure}
    \caption{Saturation in Round 1 after 160 queries, in Round 2 after 100 queries and in Round 3 after 70 queries are generated. Red star denotes satisfaction of slope condition (line 13) in Algorithm \ref{alg:sat_check}} 
    \label{fig:saturation_comparison}
\end{figure}
The backpropagation step consists of training the model to generate 1) accurate steps and, 2) Novel query leading to novel user-sensitive states. The first objective is achieved through supervised finetuning (SFT) of the model with prompt-response pairs from roll-out episodes. The second objective is achieved through Group Relative Policy Optimization (GRPO) and reward modeling. We first construct a roll-out buffer of episodes called as \textit{memory bank} in order to prepare the training data. The query generated by the explorer LM are rolled-out, stored in the \textit{memory bank} and updated iteratively (Figure \ref{fig:framework}). It consists of queries $u_i$ and steps $a_{ij}$; $i \in \{1, \dotsc I \}$ and $j \in \{ 0, \dotsc T-1\}$. Besides this, \textit{memory bank} contains text embeddings of $u_i$ and $a_{ij}$ generated by Qwen3-Embedding-0.6B Model. We design following novelty score to compose the reward:
\flushleft \textbf{Query Novelty Score}: 
    \begin{itemize}
    \item Cluster queries in memory bank and generated query with kmeans for number of clusters ranging up to total number of queries to be clustered. Find kmeans inertia for each of them
    \item Find optimal no. of cluster (= argmin of first derivative of inertia) and $same\_cluster\_indices\_query$
    \item $query\_novelty\_score= exp\{-len(queries[j]) \}$; $j \in same\_cluster\_indices\_query$
    \end{itemize}
\flushleft \textbf{Step Novelty Score}: 
    \begin{itemize}
        \item Compute the cosine similarity matrix $Sim[i, j]$ for each step `actionPurpose" with other steps in memory bank
        \item $step\_novelty\_score=
        \Sigma_i[1- max_j(Sim[i, j])]/len(steps)$; 
        $i<= len(steps), j<=len(steps\_memory\_bank) $
    \end{itemize}\flushleft \textbf{Category Novelty Score}:
    \begin{itemize}
        \item Count number of times each category has been generated in memory bank
        \item Define $c_i$ as count of category in step $i$ of generated query. $c_i=0$ if category is $not\_critic$ and $c_i=1$ if it is $critic$
        \item $category\_novelty\_score = \Sigma_i  1/len(steps)*(1+c_i)$
    \end{itemize}
The total reward for queries $u_i$ and steps $a_{ij}$ is computed as:
\begin{align}\label{eq:total_reward}
    &\mathcal{R}(u_i, a_{ij}):= \\ \nonumber
    &query\_novelty\_score(u_i)* [ step\_novelty\_score(a_{ij}) \\ \nonumber
    &+ category\_novelty\_score(a_{ij}) ]
\end{align}

\begin{figure}
    \centering
    \includegraphics[width=0.7\linewidth]{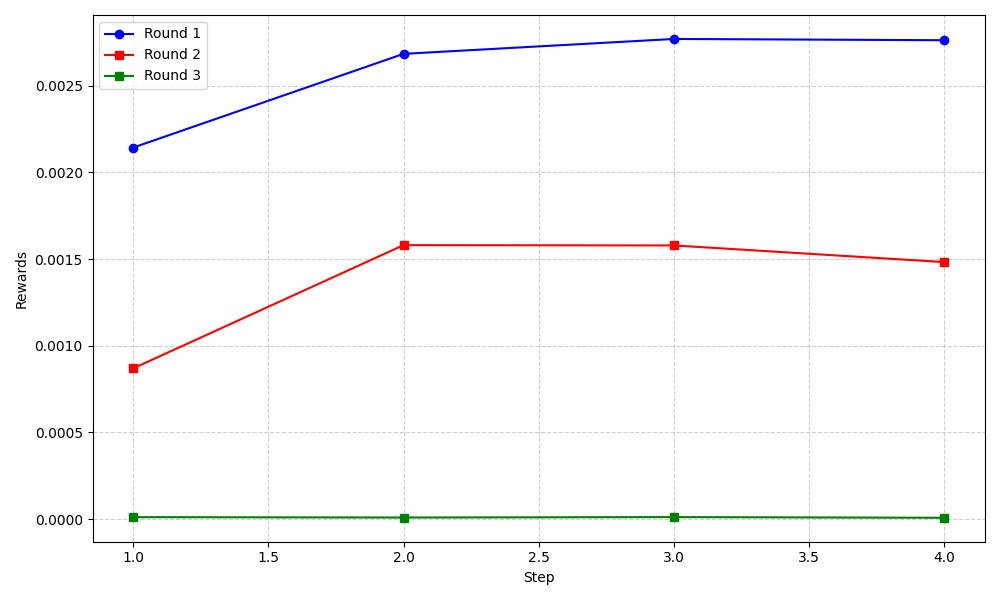}
    \caption{Total rewards over training steps across 3 training rounds}
    \label{fig:rewards_rounds}
\end{figure}
\begin{table}
    \centering
    \begin{tabular}{l c c c}
        \toprule
        \multirow{2}{*}{\textbf{Queries}} & \multicolumn{3}{c}{\textbf{Novelty Score}} \\
        \cmidrule(lr){2-4}
        \textbf{Selected} & \textbf{Query} & \textbf{Step (Predicted)} & \textbf{Step (Rolled-out)} \\
        \midrule
        13 & 0.0299 & 0.2195 & 0.1785 \\
        9  & 0.0211 & 0.2190 & 0.1635 \\
        9  & 0.0217 & 0.1790 & 0.1357 \\
        \bottomrule
    \end{tabular}
    \caption{Novelty scores across different aggregated sets in round 1}
    \label{tab:novelty_R1}
\end{table}

\begin{table}
    \centering
    \begin{tabular}{l c c c}
        \toprule
        \multirow{2}{*}{\textbf{Queries}} & \multicolumn{3}{c}{\textbf{Novelty Score}} \\
        \cmidrule(lr){2-4}
         & \textbf{Query} & \textbf{Step (Predicted)} & \textbf{Step (Rolled-out)} \\
        \midrule
        10 & 0.0123 & 0.1197 & 0.1097 \\
        20 & 0.0096 & 0.1187 & 0.1127 \\
        30 & 0.0093 & 0.1232 & 0.1129 \\
        \bottomrule
    \end{tabular}
    \caption{Novelty scores across different aggregated sets in round 2}
    \label{tab:novelty_R2}
\end{table}

\section{Experiments and Results}\label{sec:experiments}
We set  $N_{min}=70$ in Algorithm \ref{alg:learning_framework}, $\tau = 0.85$, batch size $M=10$ in Algorithm \ref{alg:query_selection}, $N_{max} = 200$ in Algorithm \ref{alg:sat_check}. Llama3.1-8B and Qwen2.5-3B-Instruct models were unable to proceed beyond initial screen and hence produced very low novelty scores in the first round. The mean query novelty score for Qwen2.5-32B-Instruct in round 1 is approximately 0.0519 compared to approximately 0.0042 for Llama 3.1 8B. We therefore choose the explorer LMs as Qwen-2.5-32B-Instruct. We performed SFT using Llama Factory with LoRA (rank=64, all linear layers) and a cosine learning rate scheduler (key hyperparameters: learning rate: 1e-4, batch size: 1, max steps: 60, infrastructure details: DeepSpeed ZeRO-3, fp16, 8 NPUs). We fine-tuned the SFT model using Group Relative Policy Optimization (GRPO) on a cluster of 4 nodes with 8 NPUs (8GB each) each, leveraging the VerL framework with reward function described in Section \ref{sec:training} and the total reward in \eqref{eq:total_reward}. Training employed a fixed learning rate of $5 \times 10^{-7}$, a batch size of 64, and 2 epochs, with model merging performed post-training to consolidate FSDP shards. We conducted 3 training rounds and observed that: 
(i) The total reward reduces progressively in orders of magnitude from $10^{-2}$ to $10^{-5}$ in subsequent training rounds (or MCTS iterations) implying a shrinking of query and screen space in each iteration (Figure \ref{fig:rewards_rounds}). This is also evidenced by progressive drop in number of queries up to saturation ($n$ in Algorithm \ref{alg:sat_check}) over subsequent training rounds as seen in Figure \ref{fig:saturation_comparison}.
(ii) The standard deviation of $step\_novelty\_score$ in Round 1 is approximately 0.0686 compared to 0.00240 and 0.0068 in Round 2 and Round 3 respectively, showing that step prediction accuracy improves due to SFT.
(iii) Instruction tuning enables novel query generation in subsequent batches within the same training round as novelty score reduces with increase of batch number (as seen in Tables \ref{tab:novelty_R1}, \ref{tab:novelty_R2}, \ref{tab:novelty_R3}). 

\begin{algorithm}[H]
\caption{Saturation Check Algorithm $\mathcal{A}_{sat}$}
\label{alg:sat_check}
\begin{algorithmic}[1]
\Require Maximum number of batches to be generated $N_{max}$, Linear fit Function $f_{linear}$, Polynomial fit Function $f_{\text{poly}}$, execution trajectory $\mathcal{T}_0$
\State Initialize batch counter $n \leftarrow 1$, Saturation Set $\mathcal{S}_{sat}=0$
\State Initialize query set $\mathcal{Q} \leftarrow \mathcal{T}_0$, batch number $x \leftarrow \emptyset$
\State Number of queries per batch $y \leftarrow \emptyset$ 
\State $\text{IsSaturated}(x, y) = \textbf{False}$
\While{$n \leq N_{max}$ \textbf{and} \textbf{not} $\text{IsSaturated}(x, y)$}
    \State Obtain $query\_retained$, $index\_retained$ from $\mathcal{A}_{sel}$ (Algorithm \ref{alg:query_selection}) with $P=Q$
    \State $\mathcal{Q} \leftarrow \mathcal{Q} \cup \{\text{query\_retained from Batch  } x\}$
    
    \State $y \leftarrow y \cup \text{len}(\text{index\_retained from Batch } x)$\;
    
    \State $x_{elbow} \leftarrow 0$, $x_{\text{coarse}} \leftarrow \text{linspace}(\min(x), \max(x), 3)$\;
    \State $y_{\text{approx}} \leftarrow f_{\text{linear}}(x, f_{\text{poly}}(y))(x_{\text{coarse}})$\;
    \State $slope_{start} \leftarrow \frac{y_{approx}[1] - y_{approx}[0]}{x_{coarse}[1] - x_{coarse}[0]}$\;
    \State $slope_{end} \leftarrow \frac{y_{approx}[2] - y_{approx}[1]}{x_{coarse}[2] - x_{coarse}[1]}$\;
    
    \If{$slope_{start} < slope_{end}$ \textbf{and} $2 \in y$}
        \State Set elbow to middle point: $x_{elbow} \leftarrow x_{coarse}[1]$\;
        \State $n_{sat}= n_{sat} + 1$, $\mathcal{S}_{sat} \cup \{n \}$
        \If{$n_{sat}==2$ \textbf{and}  $\exists k, k+1 \in  \mathcal{S}_{sat}$; $k<n-1$   }:
            \State $\text{IsSaturated}(x, y) = \textbf{True}$
        \EndIf
    \EndIf
    
    \State $n \leftarrow n + 1$, $x \leftarrow x \cup \{n\}$
\EndWhile
\State \Return $n$, $Q$
\end{algorithmic}
\end{algorithm}

\begin{table}
    \centering
    \begin{tabular}{l c c c}
        \toprule
        \multirow{2}{*}{\textbf{Queries}} & \multicolumn{3}{c}{\textbf{Novelty Score}} \\
        \cmidrule(lr){2-4}
        \textbf{Selected} & \textbf{Query} & \textbf{Step (Predicted)} & \textbf{Step (Rolled-out)} \\
        \midrule
        10 & 0.01892 & 0.1485 & 0.1056 \\
        11 & 0.00285 & 0.1154 & 0.0821 \\
        \bottomrule
    \end{tabular}
    \caption{Novelty scores across different sets in round 3}
    \label{tab:novelty_R3}
\end{table}

\section{Conclusion}
This work proposes a method for systematically and autonomously exploring the query space of a GUI application. It was demonstrated that the space of user-sensitive queries shrinks following subsequent training rounds. We hypothesize that a more aggressive search could be achieved by rejecting queries with a novelty score below the total rewards for the previous round besides the repeating ones in Algorithm \ref{alg:query_selection}. Furthermore, step-level exploration of sensitive screens from a single base query could enable a comprehensive coverage of screen space and include queries with high cosine similarity (in query) such as `turn on notification' and `turn off notification'. These topics require further investigation and are left as future work.

\subsubsection{Appendices}
\begin{figure}
    \centering
    \includegraphics[width=\linewidth, scale=1.5]{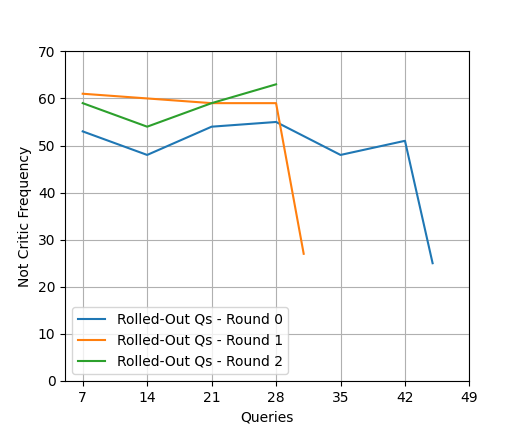}
    \caption{Frequency of `Not-Critic' screens across 3 training rounds}
    \label{fig:freq_comp}
\end{figure}

\begin{figure}
    \centering
    \includegraphics[width=\linewidth, scale=1.5]{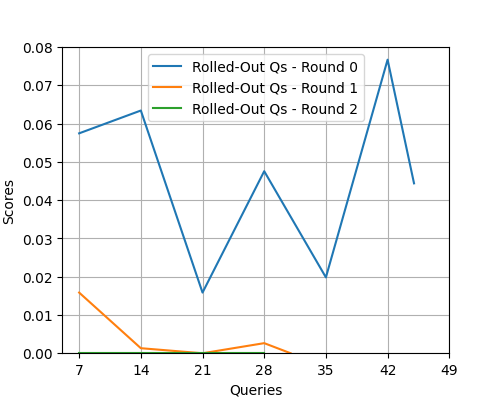}
    \caption{$category\_novelty\_score$ across 3 training rounds}
    \label{fig:score_comp}
\end{figure}

Given an example task trajectory $\mathcal{T}_0$, the instruction $\mathcal{I}$ for explorer LM is in Listing \ref{fig:proxy_prompt}. The frequency of screens which are categorized as `$not\_critic$' increases in subsequent training rounds (in Figure \ref{fig:freq_comp}) and as a result of this, the rewards for screen category for rolled out queries (in Figure \ref{fig:score_comp}) also reduces progressively. 

\begin{figure*}[t] 
    \centering
    \begin{lstlisting}[language=Python, caption={Instruction $\mathcal{I}$ used for the Explorer LM to generate new taskGoal (query) and steps}, label={fig:proxy_prompt}]
You are an assistant acting as a proxy user, and you need to create a new task description based on the example tasks provided by the user.
The task will be completed by an intelligent mobile GUI agent in a mobile environment, which will determine the specific actions for each step of the task.
Below is an example task that includes steps:
[Insert Example Task Trajectory Here]
As a proxy user, follow the steps shown above to learn about the application and create a single new task. 
For example if 'Adjust shipping preference and update contact information' is NOT a single task 
whereas 'Adjust shipping preferences for future orders' IS a single task. The app is Ebay (E-Commerce/Shopping App). 
Create diverse tasks and explore the possibilities of the application. Task with different parameters will be considered the same task: 
In the Ebay app, for example, buying a mobile phone and buying clothes are not different tasks because they both involve similar steps, 
such as opening app, searching for products, selecting specifications, and checkout. Similarly, in the Ctrip app, book hotels in Shenzhen and book hotels in Beijing. 
It's not a different mission, for the same reason. As a proxy user, your main task is to think about operational tasks that involve functions of the application.

(1) Do not generate task like those in the list: {instructions}
(2) Think of tasks which can be completed assuming the past instructions and the example tasks are already completed in the app. 
For example, if an instruction was 'add a product to the cart', you can assume the product has been added and think of critic tasks based on this information.
(3) You need to create task that may involve key operations and steps. Key steps involve authentication, personal data, 
Account management, purchasing products and privacy/location access. The categories of these screens should be in the list: {categories} 
(4) The Task Goal should not exceed 15 words and must contain the application name Ebay.
(5) You should give all the steps of the task along with uiSummary, actionPurpose, category.
    'actionPurpose' can contain a maximum of 50 words, in the same format as given in the example. 
    If there are critical actions on the screen, assign a category to them. It can be one of the following list: {categories} 
    If there is no critical action on the screen, give 'category':'not_critic'. If no suitable category is found, give 'category':'critic'. 
(6) Output a python dictionary list in the following format. 
[{ "taskGoal": "Task proposed by you as a proxy user",
"step1": {"uiSummary": "screen seen in step 1", "actionPurpose": "action along with its purpose", "category": "category, if it is a critic step; Otherwise, give not_critic"}, 
"step2": {"uiSummary": "screen seen in step 2", "actionPurpose": "action along with its purpose", "category": "category, if it is a critic step; Otherwise, give not_critic"}, 
#Add more steps as needed
"stepN": {"uiSummary": "screen seen in step N", "actionPurpose": "action along with its purpose", "category": "category, if it is a critic step; Otherwise, give not_critic"} }]
Ensure that the output complies with the Python syntax dictionary list.
    \end{lstlisting}
\end{figure*}

\appendix
\bibliographystyle{named}
\bibliography{ijcai26}

@article{Screenexplorer,
  title={State transition difference prediction for deep reinforcement learning},
  author={Chi, Haotian and Liu, Zhaogeng and Chen, Xing and Qu, Bohao and Hu, Jifeng and Jiang, Yuan and Chen, Hechang and Chang, Yi},
  journal={Pattern Recognition},
  pages={112824},
  year={2025},
  publisher={Elsevier}
}

@inproceedings{GuiBee,
  title={Gui-bee: Align gui action grounding to novel environments via autonomous exploration},
  author={Fan, Yue and Zhao, Handong and Zhang, Ruiyi and Shen, Yu and Wang, Xin Eric and Wu, Gang},
  booktitle={Proceedings of the 2025 Conference on Empirical Methods in Natural Language Processing},
  pages={33249--33266},
  year={2025}
}

@inproceedings{Guiexplorer,
  title={Gui-explorer: Autonomous exploration and mining of transition-aware knowledge for gui agent},
  author={Xie, Bin and Shao, Rui and Chen, Gongwei and Zhou, Kaiwen and Li, Yinchuan and Liu, Jie and Zhang, Min and Nie, Liqiang},
  booktitle={Proceedings of the 63rd Annual Meeting of the Association for Computational Linguistics (Volume 1: Long Papers)},
  pages={5650--5667},
  year={2025}
}

@article{Qclass,
  title={Qlass: Boosting language agent inference via q-guided stepwise search},
  author={Lin, Zongyu and Tang, Yao and Yao, Xingcheng and Yin, Da and Hu, Ziniu and Sun, Yizhou and Chang, Kai-Wei},
  journal={arXiv preprint arXiv:2502.02584},
  year={2025}
}

@inproceedings{Guiexplore,
  title={Gui-xplore: Empowering generalizable gui agents with one exploration},
  author={Sun, Yuchen and Zhao, Shanhui and Yu, Tao and Wen, Hao and Va, Samith and Xu, Mengwei and Li, Yuanchun and Zhang, Chongyang},
  booktitle={Proceedings of the computer vision and pattern recognition conference},
  pages={19477--19486},
  year={2025}
}

@article{AgentQ,
  title={Agent q: Advanced reasoning and learning for autonomous ai agents},
  author={Putta, Pranav and Mills, Edmund and Garg, Naman and Motwani, Sumeet and Finn, Chelsea and Garg, Divyansh and Rafailov, Rafael},
  journal={arXiv preprint arXiv:2408.07199},
  year={2024}
}

@inproceedings{enhancing,
  title={Enhancing decision-making for llm agents via step-level q-value models},
  author={Zhai, Yuanzhao and Yang, Tingkai and Xu, Kele and Feng, Dawei and Yang, Cheng and Ding, Bo and Wang, Huaimin},
  booktitle={Proceedings of the AAAI conference on artificial intelligence},
  volume={39},
  number={25},
  pages={27161--27169},
  year={2025}
}

@inproceedings{guardian,
  title={Guardian: A runtime framework for LLM-based UI exploration},
  author={Ran, Dezhi and Wang, Hao and Song, Zihe and Wu, Mengzhou and Cao, Yuan and Zhang, Ying and Yang, Wei and Xie, Tao},
  booktitle={proceedings of the 33rd ACM SIGSOFT international symposium on software testing and analysis},
  pages={958--970},
  year={2024}
}

@article{ReasoningLLMsare,
  title={Reasoning llms are wandering solution explorers},
  author={Lu, Jiahao and Xu, Ziwei and Kankanhalli, Mohan},
  journal={arXiv preprint arXiv:2505.20296},
  year={2025}
}

@article{Lats,
  title={Language agent tree search unifies reasoning acting and planning in language models},
  author={Zhou, Andy and Yan, Kai and Shlapentokh-Rothman, Michal and Wang, Haohan and Wang, Yu-Xiong},
  journal={arXiv preprint arXiv:2310.04406},
  year={2023}
}

@article{Nnetnav,
  title={Nnetnav: Unsupervised learning of browser agents through environment interaction in the wild},
  author={Murty, Shikhar and Zhu, Hao and Bahdanau, Dzmitry and Manning, Christopher D},
  journal={arXiv preprint arXiv:2410.02907},
  year={2024}
}

@inproceedings{Expel,
  title={Expel: Llm agents are experiential learners},
  author={Zhao, Andrew and Huang, Daniel and Xu, Quentin and Lin, Matthieu and Liu, Yong-Jin and Huang, Gao},
  booktitle={Proceedings of the AAAI Conference on Artificial Intelligence},
  volume={38},
  number={17},
  pages={19632--19642},
  year={2024}
}

@article{MCDML,
  title={Monte carlo planning with large language model for text-based game agents},
  author={Shi, Zijing and Fang, Meng and Chen, Ling},
  journal={arXiv preprint arXiv:2504.16855},
  year={2025}
}

@inproceedings{Spabench,
  title={Spa-bench: A comprehensive benchmark for smartphone agent evaluation},
  author={Chen, Jingxuan and Yuen, Derek and Xie, Bin and Yang, Yuhao and Chen, Gongwei and Wu, Zhihao and Yixing, Li and Zhou, Xurui and Liu, Weiwen and Wang, Shuai and others},
  booktitle={NeurIPS 2024 Workshop on Open-World Agents},
  year={2024}
}

@article{SeeAct,
  title={Gpt-4v (ision) is a generalist web agent, if grounded},
  author={Zheng, Boyuan and Gou, Boyu and Kil, Jihyung and Sun, Huan and Su, Yu},
  journal={arXiv preprint arXiv:2401.01614},
  year={2024}
}

@article{Androidworld,
  title={Androidworld: A dynamic benchmarking environment for autonomous agents},
  author={Rawles, Christopher and Clinckemaillie, Sarah and Chang, Yifan and Waltz, Jonathan and Lau, Gabrielle and Fair, Marybeth and Li, Alice and Bishop, William and Li, Wei and Campbell-Ajala, Folawiyo and others},
  journal={arXiv preprint arXiv:2405.14573},
  year={2024}
}

@article{Mobileagent,
  title={Mobile-agent: Autonomous multi-modal mobile device agent with visual perception},
  author={Wang, Junyang and Xu, Haiyang and Ye, Jiabo and Yan, Ming and Shen, Weizhou and Zhang, Ji and Huang, Fei and Sang, Jitao},
  journal={arXiv preprint arXiv:2401.16158},
  year={2024}
}

@article{Mobileagentv2,
  title={Mobile-agent-v2: Mobile device operation assistant with effective navigation via multi-agent collaboration},
  author={Wang, Junyang and Xu, Haiyang and Jia, Haitao and Zhang, Xi and Yan, Ming and Shen, Weizhou and Zhang, Ji and Huang, Fei and Sang, Jitao},
  journal={Advances in Neural Information Processing Systems},
  volume={37},
  pages={2686--2710},
  year={2024}
}

@article{Mobileagentv3,
  title={Mobile-agent-v3: Fundamental agents for gui automation},
  author={Ye, Jiabo and Zhang, Xi and Xu, Haiyang and Liu, Haowei and Wang, Junyang and Zhu, Zhaoqing and Zheng, Ziwei and Gao, Feiyu and Cao, Junjie and Lu, Zhengxi and others},
  journal={arXiv preprint arXiv:2508.15144},
  year={2025}
}

@article{uitars,
  title={Ui-tars: Pioneering automated gui interaction with native agents},
  author={Qin, Yujia and Ye, Yining and Fang, Junjie and Wang, Haoming and Liang, Shihao and Tian, Shizuo and Zhang, Junda and Li, Jiahao and Li, Yunxin and Huang, Shijue and others},
  journal={arXiv preprint arXiv:2501.12326},
  year={2025}
}

@article{Searchagent,
  title={Tree search for language model agents},
  author={Koh, Jing Yu and McAleer, Stephen and Fried, Daniel and Salakhutdinov, Ruslan},
  journal={arXiv preprint arXiv:2407.01476},
  year={2024}
}

@inproceedings{mobilegpt,
  title={Mobilegpt: Augmenting llm with human-like app memory for mobile task automation},
  author={Lee, Sunjae and Choi, Junyoung and Lee, Jungjae and Wasi, Munim Hasan and Choi, Hojun and Ko, Steve and Oh, Sangeun and Shin, Insik},
  booktitle={Proceedings of the 30th Annual International Conference on Mobile Computing and Networking},
  pages={1119--1133},
  year={2024}
}

@article{zhang2024privacyasst,
  title={Privacyasst: Safeguarding user privacy in tool-using large language model agents},
  author={Zhang, Xinyu and Xu, Huiyu and Ba, Zhongjie and Wang, Zhibo and Hong, Yuan and Liu, Jian and Qin, Zhan and Ren, Kui},
  journal={IEEE Transactions on Dependable and Secure Computing},
  volume={21},
  number={6},
  pages={5242--5258},
  year={2024},
  publisher={IEEE}
}

@article{he2025emerged,
  title={The emerged security and privacy of llm agent: A survey with case studies},
  author={He, Feng and Zhu, Tianqing and Ye, Dayong and Liu, Bo and Zhou, Wanlei and Yu, Philip S},
  journal={ACM Computing Surveys},
  volume={58},
  number={6},
  pages={1--36},
  year={2025},
  publisher={ACM New York, NY}
}

@inproceedings{hua2024trustagent,
  title={Trustagent: Towards safe and trustworthy llm-based agents through agent constitution},
  author={Hua, Wenyue and Yang, Xianjun and Jin, Mingyu and Li, Zelong and Cheng, Wei and Tang, Ruixiang and Zhang, Yongfeng},
  booktitle={Trustworthy Multi-modal Foundation Models and AI Agents (TiFA)},
  year={2024}
}

@article{wu2024dissecting,
  title={Dissecting adversarial robustness of multimodal lm agents},
  author={Wu, Chen Henry and Shah, Rishi and Koh, Jing Yu and Salakhutdinov, Ruslan and Fried, Daniel and Raghunathan, Aditi},
  journal={arXiv preprint arXiv:2406.12814},
  year={2024}
}

\end{document}